\documentclass[twoside,11pt]{article}

\usepackage{jmlr2e}
\usepackage{listings}
\usepackage{xcolor}

\usepackage{todonotes}

\lstdefinestyle{mystyle}{
    backgroundcolor=\color{backcolour},   
    commentstyle=\color{codegreen},
    keywordstyle=\color{magenta},
    numberstyle=\tiny\color{codegray},
    stringstyle=\color{codepurple},
    basicstyle=\ttfamily\footnotesize,
    breakatwhitespace=false,         
    breaklines=true,                 
    captionpos=b,                    
    keepspaces=true,                 
    numbers=left,                    
    numbersep=5pt,                  
    showspaces=false,                
    showstringspaces=false,
    showtabs=false,                  
    tabsize=2
}

\definecolor{codegreen}{rgb}{0,0.6,0}
\definecolor{codegray}{rgb}{0.5,0.5,0.5}
\definecolor{codepurple}{rgb}{0.58,0,0.82}
\definecolor{backcolour}{rgb}{0.95,0.95,0.92}

\newcommand{\EqualContribution}{\thanks{Equal contribution }}

\jmlrheading{1}{2000}{1-48}{4/00}{10/00}{meila00a}{Marina Meil\u{a} and Michael I. Jordan}

\ShortHeadings{Differentiable and Learnable Robot Models}{Meier, Wang, Sutanto, Lin and Shah}
\firstpageno{1}

\begin{document}

\title{Differentiable and Learnable Robot Models}

\author{\name Franziska Meier\EqualContribution \email fmeier@fb.com \\
       \addr Facebook AI Research, Menlo Park, USA
       \AND Austin Wang{\textsuperscript{*}} \email wangaustin@fb.com \\
       \addr Facebook AI Research, Menlo Park, USA
       \AND Giovanni Sutanto\thanks{The contribution was made while Giovanni Sutanto was working as an intern at Facebook AI Research in Fall 2019.} \email gsutanto@alumni.usc.edu\\
       \addr Intrinsic Innovation LLC.
       \AND Yixin Lin \email yixinlin@fb.com\\
       \addr Facebook AI Research, Menlo Park, USA
       \AND Paarth Shah \email paarth@oxfordrobotics.institute \\
       \addr Oxford Robotics Institute
}

\editor{Kevin Murphy and Bernhard Sch{\"o}lkopf}

\maketitle

\begin{abstract}
Building differentiable simulations of physical processes has recently received an increasing amount of attention. Specifically, some efforts develop differentiable robotic physics engines motivated by the computational benefits of merging rigid body simulations with modern differentiable machine learning libraries. Here, we present a library that focuses on the ability to combine data driven methods with analytical rigid body computations. More concretely, our library \emph{Differentiable Robot Models} implements both \emph{differentiable} and \emph{learnable} models of the kinematics and dynamics of robots in Pytorch. The source-code is available at \url{https://github.com/facebookresearch/differentiable-robot-model}.  
\end{abstract}


\section{Introduction}
Building or learning differentiable simulations of robots and their environments comes with several potential benefits: Massive potential for parallelization \citep{brax2021github}; enabling principled hybrid modeling via analytical models and data-driven neural networks \citep{heiden2021neuralsim, pmlr-v120-sutanto20a} and generally enabling the development of model-based reinforcement learning. The high potential of differentiable simulators has created a lot of interest and progress in developing differentiable physics engines and rigid body simulations that are differentiable \citep{brax2021github, heiden2021neuralsim, werling2021fast, de2018end, geilinger2020add,giftthaler2017automatic, hu2019difftaichi, qiao2020scalable}; some of them through analytical gradients \citep{pinocchioweb, carpentier2019pinocchio, carpentier2018analytical, de2018end}.

In this context, we introduce our library for \emph{differentiable robot models} which implements analytical \emph{differentiable} and \emph{learnable} robot models in PyTorch \citep{paszke2019pytorch}. In contrast to existing libraries, our library differs in two ways from existing libraries: 1) our library focuses on differentiable robot models (instead of full simulations with environment modeling), and provides differentiable kinematic and dynamic representations. These robot models can then, for instance, be used for control of the modeled robots in either simulation or on hardware \citep{bhardwaj2021storm, bechtle2020learning}. 2) Our library provides differentiable analytical rigid body simulations that can selectively make rigid bodies \emph{learnable}. When a rigid body becomes \emph{learnable}, we can either identify its physical constants (such as mass) from data, or learn flexible  representations (such as neural networks) for the parameters of the rigid body from data. Our learnable robot models are implemented as general torch modules, and thus can be easily used within typical deep learning pipelines.
\section{Library Overview: Differentiable and Learnable Robot Models}
\begin{figure}
    \centering
    \includegraphics[width=0.4\textwidth]{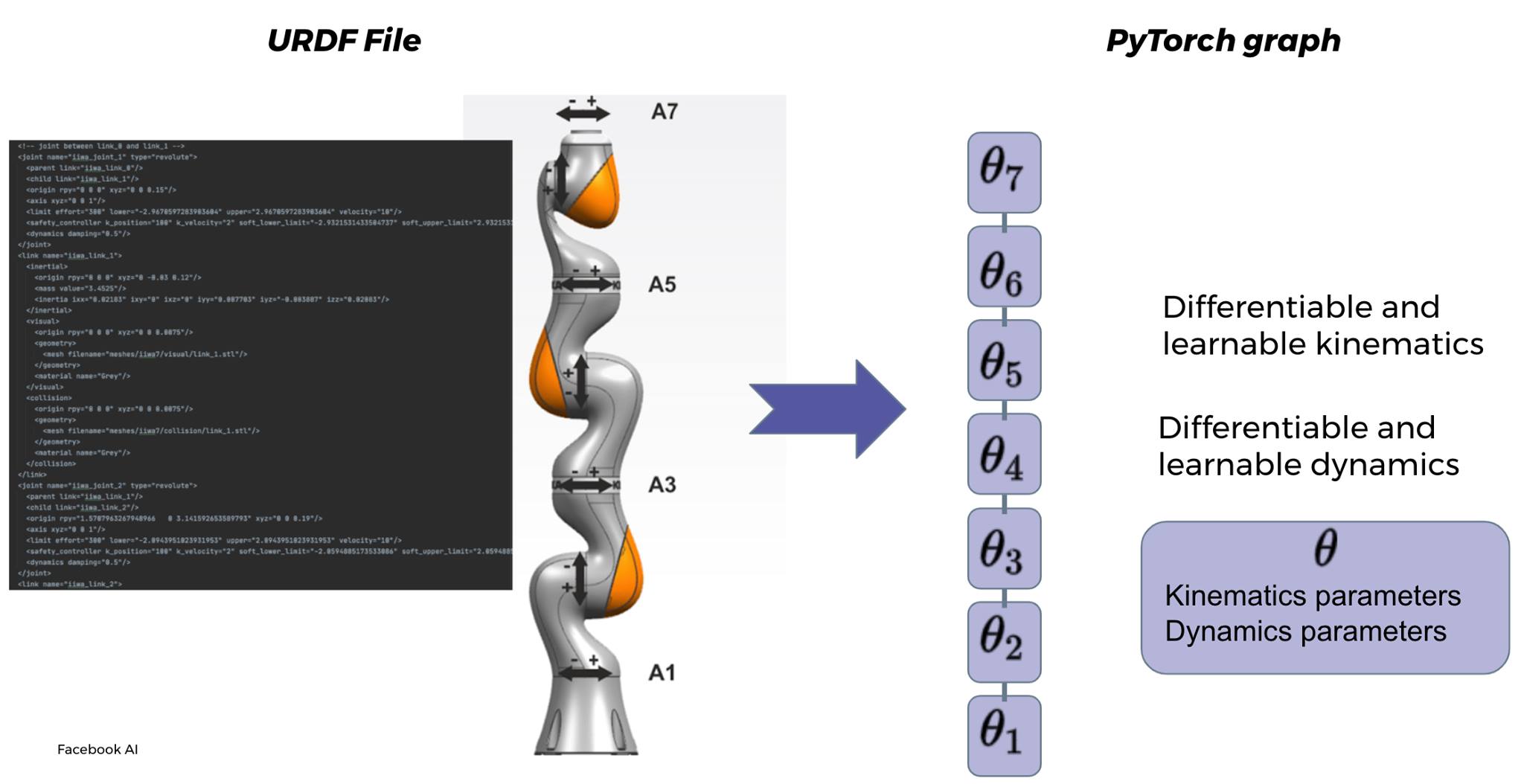}\hspace{1cm}
    \includegraphics[width=0.45\textwidth]{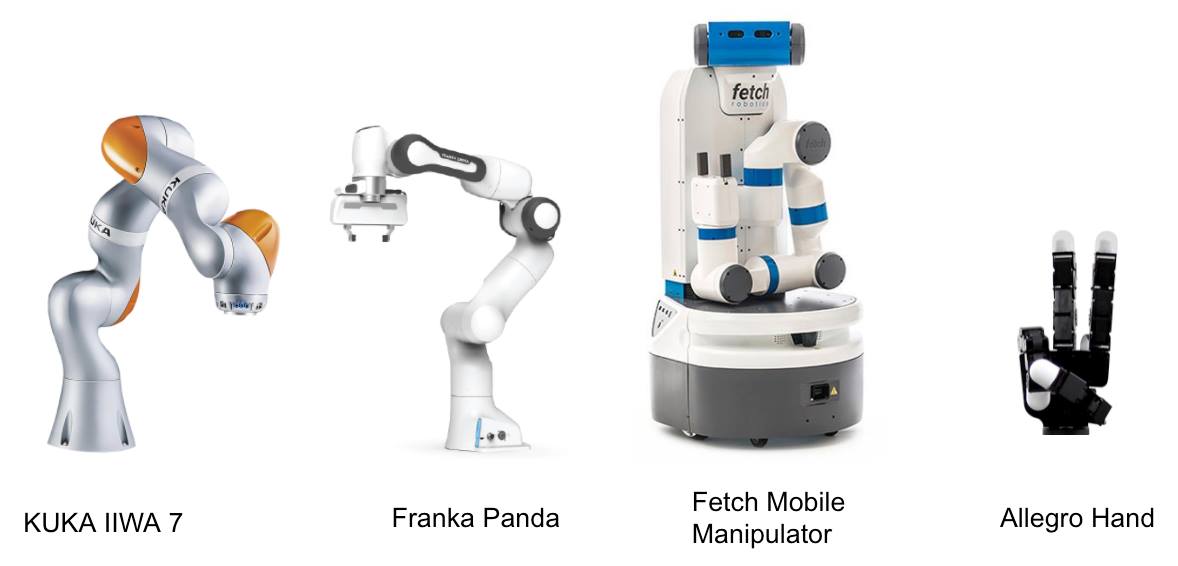}
    \caption{\small Overview: (left) Conceptual illustration of library (right) officially supported robots}
    \label{fig:library-overview}
    \vspace{-0.3cm}
\end{figure}
Our Differentiable Robot Model library parses URDFs and creates fully differentiable models. More precisely, the library constructs analytical kinematics and dynamics computations as a computation graph via PyTorch~\citep{paszke2019pytorch} for the robot described in a given URDF (see Figure~\ref{fig:library-overview}). Our library can be used on any URDF that describes a kinematic tree, and we created and tested specifically the following robots: Kuka iiwa, Franka Panda, Fetch arm and the Allegro Hand. We see at least four primary use cases for the library:
\begin{itemize}
    \item Given a complete and accurate URDF (all kinematic and dynamic parameters are accurately specified): The library can be utilized as ground truth model within model-based RL algorithms or trajectory optimization settings \citep{bhardwaj2021storm, bechtle2020learning} and enables batching and leveraging of GPU resources.
    \item Given an incomplete URDF or the goal is to study learning algorithms: We can learn any unknown parameters of the robot from observations by harnessing the power of automatic differentiation. For instance, it is possible to identify the dynamics parameters of the analytical dynamics model and compare to learning other dynamics representations with varying degrees of structure \citep{pmlr-v120-sutanto20a}.
    \item Because this library implements the analytical (ground truth) models of robots within a modern deep learning framework, it can support the pursuit of finding the "convolutional neural network" of robot motion.
    \item Finally, this library can be utilized as a teaching tool that allows to teach core robotics concepts such as kinematics in the context of modern deep learning frameworks.
\end{itemize}
The library works with any properly written URDF file that describes a kinematic tree. We officially support several robot models which are shown in Figure~\ref{fig:library-overview}. Our algorithms are carefully tested on these robots to ensure that computations within our framework are identical to that of a commercial physics engine such as Bullet \citep{coumans2016pybullet}. Our library implementation is based on spatial vector notation and implements kinematics, such as forward kinematics and Jacobian computations, as well as rigid body dynamics following Featherstone's book \citep{featherstone2014rigid}. 
Specifically, to compute the various components of the robot equations of motions, it implements the Recursive Newton Euler (RNEA) Algorithm (inverse dynamics), the Composite Rigid Body Algorithm (CRBA) (inertia matrix computation), and the Articulated Body Algorithm (ABA) (forward dynamics). 
\section{API Design and Workflow}
Our library follows the standard PyTorch API design, such that instantiating a robot model is as simple as a standard \emph{torch.nn.Module}. This design also allows for easy integration with neural-network based modules in PyTorch.
\subsection{Creating and Using a Differentiable Robot Model}
Specifically, to initialize a differentiable robot model, all the user needs is a path to a urdf:
\begin{lstlisting}[language=Python]
from differentiable_robot_model.robot_model import DifferentiableRobotModel

urdf_path = "path/to/robot/urdf"
robot = DifferentiableRobotModel(urdf_path)
\end{lstlisting}
This command will create a tree of rigid bodies, according to the structure outlined in the URDF and initialize the rigid bodies with the kinematics and dynamics parameters specified in the URDF. Once the robot model is loaded, a user can then perform various operations, such as querying the robots properties, or executing kinematic and dynamic computations:

\begin{lstlisting}[language=Python]
# Robot properties
robot.get_joint_limits()
robot.get_link_names()

# Values to query the model with
jpos = torch.rand(7), jvel = torch.rand(7), jacc_des = torch.rand(7)
torques = torch.rand(7)
ee_link_name = "iiwa_link_ee"

# Robot kinematics and dynamics
ee_pos, ee_quat = robot.compute_forward_kinematics(jpos, ee_link_name)
J_lin, J_ang = robot.compute_endeffector_jacobian(jpos, ee_link_name)

jacc = robot.compute_forward_dynamics(jpos, jvel, torques)
torques_desired = robot.compute_inverse_dynamics(jpos, jvel, jacc_des)
\end{lstlisting}

Notably, these computations are fully differentiable, and changes in the output can be propagated to the input. For example, we can compute inverse kinematics by iteratively updating the joint position until the predicted end-effector pose matches a desired pose.
\subsection{Differentiable and Learnable Rigid Body}
The key building block of the differentiable robot model is the differentiable rigid body, similar to having layers as the key building block of neural networks. The robot model is essentially a tree of connected rigid bodies where the tree connections define how inputs/outputs/gradients are propagated. Each rigid body has variables that capture the (local) kinematic and dynamic state of the body. In some applications, it might be desirable to learn parameters of the robot model. In that case, the user can make the parameters of links learnable:
\begin{lstlisting}[language=Python]
from differentiable_robot_model.rigid_body_params import PositiveScalar

# make a parameter of a link learnable
robot_model.make_link_param_learnable(<link_name>, <param_name>, <module>)
# for instance:
robot_model.make_link_param_learnable("link_1", "mass", PositiveScalar())
\end{lstlisting}
%
While our library provides implementations for parametrizations (such as a \emph{PositiveScalar} "net"), users can specify their own flexible parametrizations with a torch.nn.Module.
\subsection{Learning Loop}
Finally, to learn parameters from data, the user implements a typical learning loop
\begin{lstlisting}[language=Python]
 """ generate training data via ground truth model """
train_data = generate_sine_motion_forward_dynamics_data(gt_robot_model, n_data=n_data, dt=1.0 / 250.0, freq=0.1)
train_loader = DataLoader(dataset=train_data, batch_size=100, shuffle=False)

""" optimize learnable params """
optimizer = torch.optim.Adam(robot_model.parameters(), lr=1e-2)
loss_fn = NMSELoss(train_data.var())
for i in range(n_epochs):
    losses = []
    for batch_idx, batch_data in enumerate(train_loader):
        q, qd, qdd, tau = batch_data
        optimizer.zero_grad()
        qdd_pred = robot_model.compute_forward_dynamics(
            q=q, qd=qd, f=tau, include_gravity=True, use_damping=True)
        loss = loss_fn(qdd_pred, qdd)
        loss.backward()
        optimizer.step()
\end{lstlisting}

\subsection{Conclusion}
We present our library on differentiable and learnable robot models, that provides a flexible implementation of analytical rigid body simulations of popular robots in Pytorch. Our library works for arbitrary kinematic trees, but officially supports (with testing) a variety of robots (see Figure~\ref{fig:library-overview}). Finally, our library enables the user to selectively make rigid bodies learnable, and explore the option of identifying parameters or more complex models from data.



\vskip 0.2in
\bibliography{diff}

\end{document}